\documentclass[runningheads]{llncs}

\usepackage{eccv}

\usepackage{eccvabbrv}

\usepackage{graphicx}
\usepackage{booktabs}

\usepackage[accsupp]{axessibility}  

\newcommand{\Mmat}[0]{{{\boldsymbol M}}}

\newcommand{\Umat}{{{\boldsymbol U}}}

\newcommand{\Xmat}{{\boldsymbol X}}
\newcommand{\Ymat}[0]{{{\boldsymbol Y}}}

\newcommand{\gv}[0]{{\boldsymbol{g}}}

\newcommand{\nv}{\boldsymbol{n}}

\newcommand{\sv}[0]{{\boldsymbol{s}}}

\newcommand{\xv}{\boldsymbol{x}}
\newcommand{\yv}{\boldsymbol{y}}

\newcommand{\zv}{\boldsymbol{z}}

\usepackage{amssymb} 

\usepackage{hyperref}

\usepackage{orcidlink}
\usepackage{multirow}

\begin{document}

\title{Deep Probabilistic Unfolding for Quantized Compressive Sensing} 

\titlerunning{Deep Probabilistic Unfolding for Quantized Compressive Sensing}

\author{Gang QU\inst{1} \and
Ping Wang\inst{1} \and
Siming Zheng\inst{2} \and
Xin Yuan\inst{1}*}
\authorrunning{Qu. et al.}
\institute{Westlake University, School of Engineering, Hangzhou, Zhejiang, China \and
Vivo Mobile Communication Co., Ltd., Hangzhou, Zhejiang, China.
\email{qugang@westlake.edu.cn}\\
}


\maketitle

\begin{abstract}
  We propose a deep probabilistic unfolding model to address the classical quantized compressive sensing problem that leverages an unfolding framework to enhance the reconstruction accuracy and efficiency. Unlike previous unfolding methods that apply $L_2$ projection to measurements, we derive a closed-form, numerically stable likelihood gradient projection, which allows the model to respect the true quantization physics, turning the hard quantization constraint into a soft probabilistic guidance.
Furthermore, an efficient, dual-domain Mamba module is specifically designed to dynamically capture and fuse the multi-scale local and global features, ensuring the interactions between the distant but correlated regions. 
Extensive experiments demonstrate the state-of-the-art performance of the proposed method over previous works, which is capable of promoting the application of quantized compressive sensing in real life.
\end{abstract}

\section{Introduction}
\label{sec:intro}

Sensing is essential to human perception and understanding of the physical world, yet obtaining high-throughput data that accurately capture optical signals remains a fundamental challenge in imaging.
Compressive sensing (CS) addresses this challenge by leveraging the compressibility of natural signals.
The core principle of CS lies in acquiring significantly fewer samples than that required by the Nyquist-Shannon sampling theorem through compressed measurements, while still preserving sufficient information for accurate signal reconstruction.
This departures from conventional sampling paradigms has broad implications for various applications, including single-pixel cameras~\cite{duarte2008single}, lensless imaging~\cite{Yuan18OE, slope}, hyperspectral imaging~\cite{9363502, qin2025detail,wang2025spectral}, high-speed imaging~\cite{wang2023full,wang2023deep,wang2024hierarchical} and so on. 
The success of CS, coupled with the fundamental role of quantization in signal digitization, has fueled a growing interest in quantized CS (QCS)~\cite{zymnis2009compressed,yang2013variational,boufounos2015quantization, meng2023qcs, oh2022communication, meng2024qcs, yang2025enhancing}.
Coarse quantization is particularly
appealing as it results in significant reduction in bandwidth requirements and power consumption. 
Especially low bit quantization(1,2,3-bit), wherein the measurements are non-linearly mapped to a certain range of quantized codewords, and then the. From the information security perspective, the quantization operations are also essential to protect the privacy during the information transmission in channel. However, low-bit quantization loses information about signal amplitude, and it is impossible to recover the amplitude during signal reconstruction.
Therefore, low bit CS finds applications in systems that require the recovery of the unknown signal
up to a scaling factor. For example, in a frequency division
duplex massive MIMO system, the direction of the channel
state information at the transmitter is sufficient for the design
of beam-forming vectors. In this case, using 1-bit CS saves
the bandwidth resources required for the channel state
information feedback~\cite{tang2017low}. Some other applications where
1-bit CS is used are radar~\cite{ameri2019one}, source localization~\cite{shen2013one},
and wireless sensor networks~\cite{cao2016implementation}.

QCS was originally introduced in~\cite{boufounos20081}, and several
reconstruction algorithms have since then been proposed in the
following works. Though QCS has shown promising inference and signal reconstruction
performance, it is also known to be quite sensitive to noise~\cite{yan2012robust, li2014robust, kafle2022noisy}. Some recent works have dealt with the problem by
mitigating noise~\cite{kamilov2012one, plan2012robust, musa2016generalized}, using multiple measurement
vectors~\cite{kafle2018joint}, or by using side-information~\cite{kafle2022noisy}. These algorithms fall into the category of “traditional” algorithms as they
are model-driven, where recovery performance depends on
how well the model represents the actual sparse structure of the
signal. Among model-driven algorithms, Bayesian algorithms
often perform better than non-Bayesian counterparts due to
their ability to incorporate the prior on the sparse signal
structure through a probability distribution. 

Recently, some works also explore the possibility of using a deep learning-based approach
for QCS, i.e., using generative model to recover the signal from quantized CS measurement~\cite{meng2023qcs, meng2024qcs, kafle2025one}. 
However, these methods formulate QCS recovery as posterior sampling with score-based generative priors, where reconstruction requires iterating over multiple noise levels and performing multiple stochastic updates per level via annealed Langevin dynamics. This implies that inference is not a single feed-forward pass, but a computationally expensive iterative procedure whose runtime scales with the total number of sampling steps.
From a practical standpoint, this inference paradigm raises clear concerns for real-world deployment, where latency and throughput are difficult to guarantee. The computational cost is high, and the stochastic nature of sampling can introduce output variability unless multiple samples are averaged (which further increases cost). 
On the other hand, these methods critically depend on a pre-trained score-based generative model as the prior term in posterior sampling. While this choice provides a powerful implicit prior and yields impressive perceptual quality, it also imposes substantial training and data requirements. This dependence raises several concerns: (i) training the prior can require large-scale datasets and compute, (ii) the learned prior may not match domain-specific measurement settings, and (iii) the end-to-end sensing pipeline (including quantization scale, dithering, and measurement statistics) is not directly optimized jointly with the prior in these sampling-based frameworks.

Bearing these concerns in mind, in this paper, we propose a Deep Probabilistic Unfolding Network (DPUNet) for QCS reconstruction, where we derive a closed-form likelihood gradient projection to confine the optimization of network, and also propose an efficient, dual-domain Mamba block (DMB) as the refinement network for global feature extraction and fusion in DUN model. The main contributions of this paper are summarized as follows:
\begin{itemize}
\item We propose a DPUNet for QCS reconstruction, which is a closed-form, numerically stable likelihood data-consistency gradient projection, and allows the model to turn the hard quantization constraint into a soft probabilistic guidance. The end-to-end training strategy also makes the model focus more on the fidelity of reconstruction compared to previous methods, which is quite crucial for the downstream application scenarios.
\item A dual-domain Mamba block is specially designed in DPUNet to dynamically capture and fuse local and global features in the spatial and spectral domains. To tackle the limitation of scan-based state space mixers of their inherent order dependence, we parameterize a frequency-conditioned, stable complex response, yielding an order-invariant global receptive field in Fourier spectrum. Furthermore, to capture departures from strict diagonal structure without incurring dense spectral mixing, we further augment the diagonal operator with a structured low-rank projection–broadcast term, enabling efficient cross-frequency interactions at low computational cost. This design provides a principled alternative to multi-direction scanning, which means global information modulation and interaction are achieved through structured spectral operators rather than through repeated order-dependent recurrences.
\item The extensive results on different datasets demonstrate the feasibility and effectiveness of the proposed method compared to prior works in both reconstruction accuracy and efficiency. The proposed method pave the way for QCS reconstruction in a more efficient and reasonable end-to-end design in future works.
\end{itemize}

\section{Related Works}
\subsection{Quantized Image Compressive Sensing}
Image CS reconstruction methods can be classified into two categories: model-based methods~\cite{kim2010compressed,dong2014compressive,Metzler2016FromDT} and learning-based methods~\cite{zhang2018ista,shen2022transcs, zheng2024block, Wang_2025_CVPR, 11251252}. The conventional model-based methods mainly rely on some hand-crafted priors to recover the original image from its sub-sampling measurement in an iterative manner, which 
enjoy high generalization and robustness, but meanwhile suffer from high computational cost and limited reconstruction quality. 
Earlier deep learning-based methods~\cite{kulkarni2016reconnet,metzler2017learned, qu2022demosaicing} treat DNN as a black box and  directly build a mapping from compressed measurement to the image. Recently, DUNs are proposed to incorporate DNN with conventional model-based methods, and train the unfolding network with multiple stages in an end-to-end manner, which enjoys good interpretability and has become the mainstream for CS reconstruction. Different optimization methods lead to different optimization-inspired DUNs, e.g., proximal gradient descent (PGD) algorithms~\cite{chen2022content, song2021memory}, AMP~\cite{zhu2020deformable}, ADMM~\cite{Wang_2025_CVPR} and so on. Although DUN presents superiority compared to the end-to-end design, the introduction of gradient projection inevitably increases the computational burden, especially for large-scale image with its corresponding large sensing matrix. Thus, Block-based~\cite{zhang2018ista,shen2022transcs} and Kronecker-based~\cite{qu2024dual,  Wang_2025_CVPR} DUNs have been proposed for both efficient and high-quality CS reconstruction in previous works.

The introduction of quantization operation in CS problems leads to QCS~\cite{zymnis2009compressed}, the forward model can be expressed as:
\begin{equation}
\label{eq:QCS}
\yv = Q(\Mmat\xv + \nv).
\end{equation}
The inverse problem aims to recover the original signal $\xv \in \mathbb{R}^{N\times 1}$ from the quantized compressed measurement $\yv \in \mathbb{R}^{M\times 1}$ with the sensing matrix $\Mmat \in \mathbb{R}^{M \times N}$, $\nv$ denotes an additive noise following i.i.d. and gaussian distribution. $Q(\cdot)$ denotes element-wise quantization function that maps each element to a set of codewords $\Im$, i.e., $\yv_i =  Q(z_i + n_i) \in \Im$, where $z_i$ is the $i$-th element of $\zv = \Mmat\xv$. For 1-bit case, i.e., $Q=1$, $Sign$ function is employed to obtain the binarized measurement:
$\yv = Sign (\Mmat\xv + \nv)$,
thus the quantization codewords are $\Im = \left \{-1, 1 \right \}$.
For multi-bit cases, the quantization codewords usually correspond to intervals. Consider a uniform quantizer with $Q$ quantization bits, the quantization codewords $\Im =\left\{q_r\right \}_{r=1}^{2^Q}$ consist of $2^Q$ elements, where $q_r = \frac{(2r-2^Q-1)\Delta}{2}$, $r\in [1, 2^Q]$, $\Delta>0$ denotes the quantization interval. 


\subsection{QCS Reconstruction with Generative Model}
Quantization leads to severe information loss at low quantization bit, which makes the CS recovery particularly
challenging.
Generative models have been adopted for low-bit QCS reconstruction in previous works~\cite{meng2023qcs, meng2024qcs, kafle2025one}. If the score function $\nabla \log p(\rm x)$ for a continuous differentiable probability density function $p(\rm x)$ is available, then we can iteratively sample from it using Langevin dynamics:
\begin{align} \label{eq:langevin}
{\rm x}_t = {\rm x}_{t-1} + \alpha_t\nabla {\log} \, p({\rm x}_{t-1}) + \sqrt{2\alpha{_t}} \, z_t, 
\end{align}
where $t \in [1, T]$, $T$ is the total number of iterations, $\alpha_t >0$ denotes the step size, and $\zv\sim {\cal N}(z_t; \,0, \rm I)$. The score function for a certain distribution can be estimated using a score network $\rm s_\theta (x)$ via score matching~\cite{vincent2011connection}. An annealed version of Langevin dynamic was proposed in~\cite{song2019generative}, which perturbs the data with Gaussian noise of different scales and jointly estimates the score functions of noise-perturbed data distributions.  The noise conditional score network (NCSN) $\rm s_\theta (x, \beta)$ aims to estimate the score function of each $p_{\beta_t}(\rm \tilde x)$ by optimizing the following weighted sum of score matching objective:
\begin{align} \label{eq:NSCN}
&\theta^* = \underset{\theta}{\arg\min} \sum_{t=1}^{T} \mathbb{E}_{{p_{data}}(\rm x)}\mathbb{E}_{p_{\beta_t}(\rm \tilde x|x)}[\left \| s_\theta(\tilde x, \beta_t) - \nabla_{\tilde x} \log p_{\beta_t}(\tilde x|x)   \right \|^2_2].
\end{align}
After training the NCSN, we can perform $K$ steps of Langevin MCMC to obtain a sample for each $p_{\beta_t}(\tilde x)$ for each noise scale:
\begin{align} \label{eq:langevin2}
{x}_t^k = {x}_{t}^{k-1} + \alpha_t{s_{\theta}}({x}_{t}^{k-1}, \beta_t) + \sqrt{2\alpha{_t}}, z_t^k, 
\end{align}
where $k \in [1,K]$. The sampling process is repeated for $t=1,2,...,T$ sequentially with $\rm x_1^0 \sim \cal N \rm (x;0,\beta_{max}^2I)$ and $x^0_{t+1}=x^K_t$ when $t<T$. When $K\rightarrow \infty$ and $\alpha\rightarrow 0$ for all t, the last sample $x^K_T$ will be a precise sample from $p_{\beta_{min}}(\tilde x) \approx p_{data}(x)$ under some regularity conditions.

In quantized CS with known binarized compressed measurement $\Ymat$, we aim to sample from the posterior distribution $p(\rm x|\,y)$ rather than $p(\rm x)$. Thus, the sampling process in Eq.~\eqref{eq:langevin} turns out to be:
\begin{align} \label{eq:pc}
{\rm x}_t = {\rm x}_{t-1} + \alpha_t\nabla {\rm x}_{t-1} {\log} \, p({\rm x}_{t-1}|y) + \sqrt{2\alpha{_t}} \, z_t. 
\end{align}
In QCS-SGM~\cite{meng2023qcs}, a noise-perturbed pseudo-likelihood score is derived to match the diffusion perturbation, but the tractability and accuracy of this likelihood approximation rely on structural properties of the sensing operator, particularly assumptions closely related to row-orthogonality. When these assumptions are violated (e.g., correlated or ill-conditioned sensing matrices), the pseudo-likelihood approximation can degrade and lead to unstable or inaccurate gradients. QCS-SGM+~\cite{meng2024qcs} is explicitly motivated by this limitation and replaces the closed-form approximation with an EP-based approximation to handle general matrices. However, this improvement comes at the cost of a heavier inference procedure and additional algorithmic complexity, making it especially problematic for time-sensitive applications.

\section{Proposed Methods}
\begin{figure}[tb]
  \centering
  \includegraphics[width=.9\linewidth]{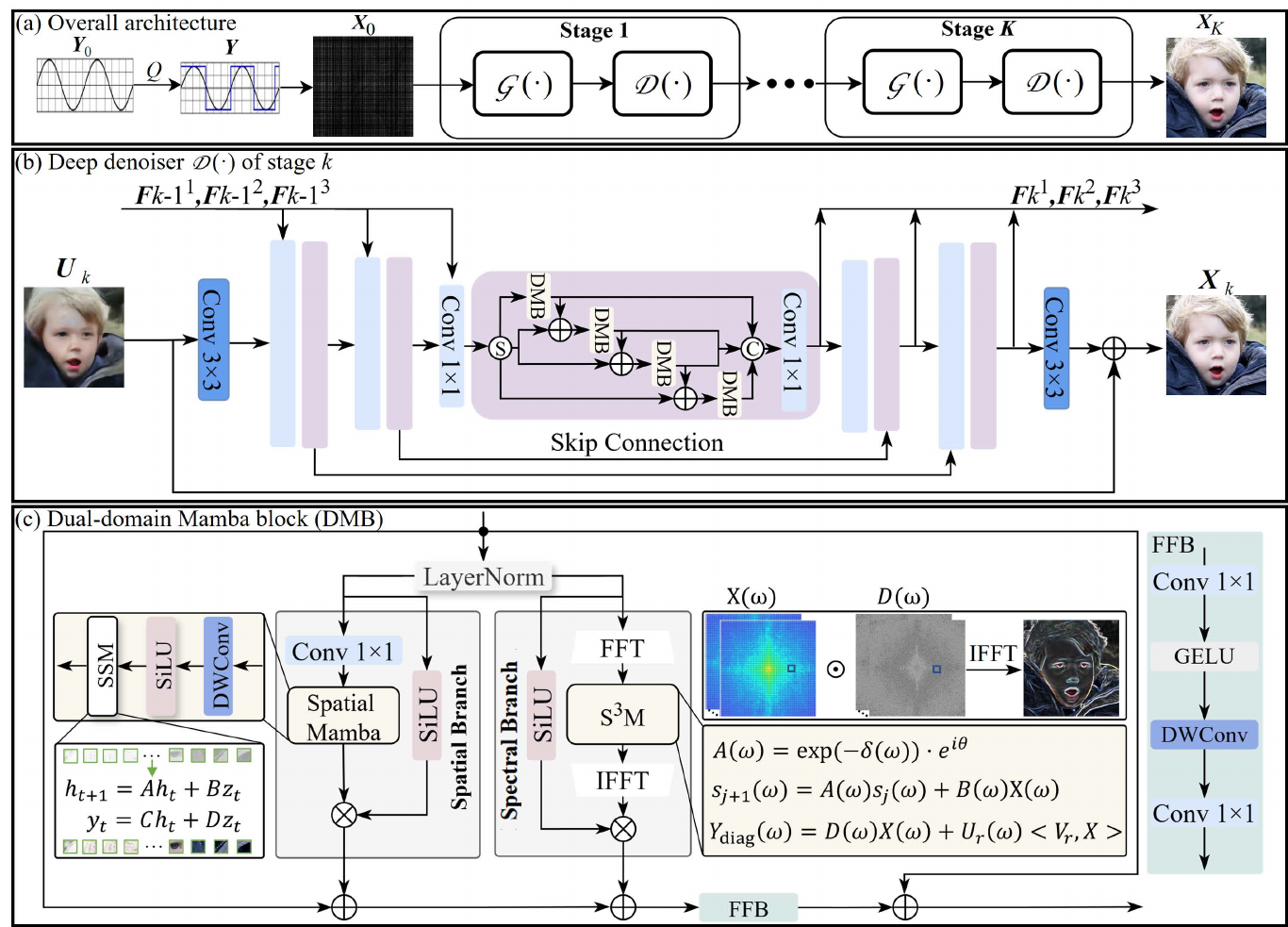}
  \caption{The pipline of the proposed DPUNet. (a) The overall architecture of the proposed QCS reconstruction model, which is a DUN framework with $K$ stages. (b) The design of deep denoiser in (a). (c) The proposed dual-domain Mamba block in (b).
  }
  \label{fig:overallmodel}
\end{figure}



\subsection{Probabilistic Unfolding Network for QCS}
We propose deep probabilistic unfolding network (DPUNet), an end-to-end trainable unfolding framework for QCS reconstruction. In contrast to QCS-SGM~\cite{meng2023qcs} and QCS-SGM+~\cite{meng2024qcs}, which perform posterior sampling using a pre-trained score model, DPUNet compiles the Bayesian inference updates into a deterministic, multi-stage network. Each stage consists of (i) an explicit noise-perturbed quantized data-consistency likelihood gradient projection and (ii) a learnable refinement block. The goal is to preserve principled data-consistency from quantized likelihoods while avoiding expensive posterior sampling and reducing reliance on restrictive operator assumptions.
\subsubsection{Likelihood Gradient Projection.}
Our method explicitly incorporates a closed-form quantized likelihood data-consistency projection: we use Mills-ratio-based expressions for the 1-bit case and interval-based Gaussian cumulative distribution function (CDF) differences for multi-bit quantization. This retains a clear probabilistic grounding (data-consistent updates) while enabling end-to-end learning of the update strength and schedule. In contrast to EP-based likelihood approximations~\cite{meng2024qcs}, the resulting update is lightweight and amenable to GPU-accelerated feed-forward execution.
In DPUNet, we do not simply rely on hard quantization. Instead, we assume the system is perturbed by noise, either inherent physical noise or artificially added smoothing noise. 
At stage $k$, the current estimate $\xv_k$ induces $\zv=\Mmat\xv_k$, $M$ denotes the sensing matrix. We model the effective measurement-domain standard deviation as:
\begin{align} \label{eq:Standarddeviation}
\epsilon_k = \sqrt{\sigma^2+\beta^2_kd},
\end{align}
where $\beta_k$ is learned stage noise level, $d$ is the diagonal of $\Mmat\Mmat^\top$. The noise-perturbed quantized likelihood is then evaluated under Gaussian perturbation with scale $\epsilon_k$.
DPUNet considers an explicit log-likelihood gradient with respect to $z_k\sim\mathcal{N}(z, \epsilon_k^2)$:
\begin{align} \label{eq:explicitgradient}
\gv_k^{(z)}\triangleq \nabla_{\zv} {\rm log}p(y|\zv; \beta_k, \sigma^2),
\end{align}

For $Q>1$, $y$ corresponds to interval constraints $z_k \in [y_{l}, y_{u}]$, observing a quantized value $y_k$ implies that the latent non-quantized value $v_k$ falls within the specific interval.  We need to calculate the probability of observing $\yv$ given the current estimate $\xv$. Since the effective noise is gaussian, the likelihood is the integral of the Gaussian probability distribution function (PDF) over the quantization interval:
\begin{align} \label{eq:likelihood}
p(y|\zv_k) = \int_{y_{l}}^{y_{u}} \mathcal{N}(t;\zv_k, \epsilon_k^2) dt.
\end{align}
Using the CDF of the standard normal distribution, denoted as $\varphi(\cdot)$, we can standardize the integral limits:
\begin{align} \label{eq:cfd}
p(y|\zv_k) = \varphi(\frac{y_{u}-\zv_k}{\epsilon_k}) - \varphi(\frac{y_{l}-\zv_k}{\epsilon_k}).
\end{align}
Then differentiate the log-likelihood:
\begin{align} \label{eq:likelihoodGD}
\nabla_{\zv_k} {\rm log} p(y|\zv_k) = \frac{1}{\varphi(\tilde{u}_k) - \varphi(\tilde{l}_k)}(\phi(\tilde{u}_k)\frac{\partial\tilde{u}_k}{\partial z_k}-\phi(\tilde{l}_k)\frac{\partial\tilde{l}_k}{\partial z_k})= \frac{\phi(\tilde{l}_k) - \phi(\tilde{u}_k)}{\varphi(\tilde{u}_k) - \varphi(\tilde{l}_k)}\cdot \frac{1}{\epsilon_k},
\end{align}
where $\phi(\cdot)$ denotes the standard normal PDF.

For the 1-bit case, $y\in [-1, 1]$, the observation corresponds to a half-space event:
\begin{align} \label{eq:1bitcase}
y=+1 \longleftrightarrow {z_k} \geq0, y=-1 \longleftrightarrow {z_k} \leq0.
\end{align}
Therefore, we have:
\begin{align} \label{eq:1bitcasep}
p(y|z_k;\epsilon^2) = \mathbb{P}(\mathcal{N}(z_k, \epsilon^2)\geq0)^{\mathbb{I}[y=+1]}\mathbb{P}(\mathcal{N}(z_k, \epsilon^2)\leq0)^{\mathbb{I}[y=-1]}=\varphi(\frac{yz_k}{\epsilon}).
\end{align}
Similarly, take the log and differentiate:
\begin{align} \label{eq:1bitcasep}
\frac{\partial}{\partial z}{\log } p(y|z_k;\epsilon^2) = \frac{\partial}{\partial z} {\log} \varphi(\frac{yz_k}{\epsilon})=\frac{y}{\epsilon}\cdot\frac{\phi(\frac{yz_k}{\epsilon})}{\varphi(\frac{yz_k}{\epsilon})}.
\end{align}
The ratio $\mathcal{M}(t)\triangleq \frac{\phi(t)}{\varphi(t)}$ is the Mills ratio, hence:
\begin{align} \label{eq:MillsRatio}
\nabla_{\zv}{\rm log}p(y|z_k) = \frac{y}{\epsilon_k}\mathcal{M}(\frac{yz_k}{\epsilon}).
\end{align}
After mapping the measurement-domain score to the image domain, the likelihood gradient descent projection in stage $k$ can be expressed as:
\begin{align} \label{eq:NSCN}
{\mu}_{(k+1)}  & = {x}_{(k)} + \lambda_k\nabla_{{\zv}}\log p(y|{x}_k) \\ \notag &={x}_{(k)} + \lambda_k(M^\top \gv_k^{(z)}),
\end{align}
where $\lambda_k$ is a learnable step size, $M$ and $M^\top$ denote the sensing matrix and its transpose.

From the above deduction, our method does not require row-orthogonality as a principled prerequisite for inference, as we perform deterministic deep unfolding with a fixed number of stages, rather than iterative posterior sampling. Nonetheless, to obtain a lightweight and numerically stable data-consistency update, our implementation adopts a diagonal noise approximation in the measurement domain and parameterizes the effective perturbation scale as Eq.~\eqref{eq:Standarddeviation}, which yields closed-form 1-bit Mills-ratio and multi-bit Gaussian CDF difference likelihood gradients. In the case where $MM^\top$ is not diagonal, the unfolding model is capable of adjusting the mismatch by learning through stage-wise schedules, calibrated likelihood updates, and learned refinement blocks.
\subsection{Design of Deep Denoiser}
\subsubsection{Overall Architecture.} 
As shown in Fig.~\ref{fig:overallmodel} (b), the proposed deep denoiser ${\mathcal D(\cdot)}$ is basically a symmetric UNet architecture for multi-scale representation learning, composed of two encoder layers, a bottleneck layer, and two decoder layers. 
The input of the deep denoiser is the result of the likelihood gradient projection $\Umat_{k}$, which then passes through encoders to generate the deep features. In each encoder layer, the features of current layer are concatenated with features from previous stage and fused by channel concatenation and $1\times1$ convolution. To reduce the computational complexity, we separate the input features into four groups along the channel dimension and adopt the residual connection for interactions among stages. The encoder in each layer mainly contains dual-domain Mamba block (DMB) and the down-sampling layer. The goal of encoders is to progressively reduce the spatial resolution by half and double the channel dimensions, yielding the multi-scale features transferred to the decoder by skip connections. In the decoder branch, the same design with encoder is adopted, with simple $1\times1$ convolution and pixel shuffle operations for inner-stage feature fusion and up-sampling. The output of each decoder is the feature map of the current stage and are then transferred to the next stage. Finally, the feature map from the last decoder is converted to the output $\Xmat_{K}$ of current stage with a $3\times3$ convolution.
\subsubsection{Dual-domain Mamba Block}
In this section, we introduce dual-domain Mamba block that contains spatial and Spectral State Space Modeling ($S^3M$). $S^3M$ is a complex-valued, frequency-aligned spectral layer that combines stable state-space dynamics with explicit spectral filtering and efficient cross-frequency interactions through a structured Diagonal and Low-rank operator.

{\noindent \bf Spatial branch.}
We first use the Layernorm to process the input feature from $F_{in}$ to $F_{LN}$. Then,
the spatial branch operates on a flattened spatial sequence 
$\zv \in \mathbb{R}^{B\times N\times C}$ 
with $N=HW$, updating a hidden state along the token index
$t\in \{ 1,...,N \}$:
\begin{align} \label{eq:SpatialM}
h_{t+1}=Ah_t+Bz_t, y_t=Ch_t+Dz_t,
\end{align}
where $h_t$ is a latent state and ($A,B,C,D$) are learned parameters. The output of spatial branch is $Y_{spa}=y_N\odot {\rm SiLU}(F_{LN})$.

{\noindent \bf Spectral branch.}
The spectral branch is built on the classical fact that translation-equivariant linear operators are diagonalized by the Fourier basis, as shown in Fig.~\ref{fig:overallmodel} (c). Given an input feature map ${\rm X_{in}} \in \mathbb{R}^{B\times C\times H\times W}$, we first compute its half-spectrum via the real 2D Fourier transform:
\begin{align} \label{eq:rFFT2}
\Xmat_{f} = {\rm rFFT2}({\rm X_{in}})\in \mathbb{C}^{B\times C\times H\times W_f},
\end{align}
where $\rm rFFT2(\cdot)$ is the real-to-complex 2D Fourier transform storing only the non-redundant half spectrum, $ W_f=[W/2]+1$ is the retained half-spectrum width. For each frequency bin $\omega$ (an index over the spectral grid $H\times W_f$), the slice $\Xmat_{f}(\omega)\in \mathbb{C}^{B\times C}$ contains complex Fourier coefficients for all batch elements and channels.
The spectral branch defines a frequency-conditioned complex state-space recurrence. For each frequency bin $\omega$,
we define a latent state sequence $\{ \sv_j(\omega)\}_{j=0}^J$, where $J\in\mathbb{N}$ is the number of recurrent steps, then $S^3M$ instantiates a complex-valued driven state-space system:
\begin{align} \label{eq:State-spaceSystem}
&s_{j+1}(\omega) = A(\omega)s_{j}(\omega) + B(\omega)X(\omega), 
\end{align}
with $s_0(\omega)=0$, and the spectral output is expressed as:
\begin{align} \label{eq:out}
&Y_{\rm diag}(\omega) = C(\omega)s_J(\omega),
\end{align}
 We enforce stability by parameterizing the transition as follows:
\begin{align} \label{parameterizing}
A(\omega) = \rm exp(-\delta(\omega))\cdot e^{i\theta (\omega)}, \delta(\omega
)\geq0, \theta(\omega) \in (-\pi, \pi),
\end{align}
Unlike spatial scanning where $y_t$ must be produced by sequential recurrence, the above driven recurrence admits a closed-form solution:
\begin{align} \label{closedform}
s_J(\omega)=(\frac{1-A(\omega)^J}{1-A(\omega)})B(\omega)X(\omega).
\end{align}
Then, the spectral response becomes an explicit complex filter:
\begin{align} \label{complexfilter}
Y_{\rm diag}(\omega) = D(\omega)X(\omega), D(\omega)=C(\omega)B(\omega)(\frac{1-A(\omega)^J}{1-A(\omega)}).
\end{align}
Thus, rather than scanning in frequency domain, the proposed $S^3M$ adopts a recurrence over state steps $J$ within each frequency eigenmode $\omega$, whose result is a frequency response $D(\omega)$. This is Mamba-style in the sense that it is induced by a stable SSM parameterization, which is not a token scan but a global perception comes from operating on Fourier modes, each of which is spatially global.

Pure diagonal filtering assumes that each frequency bin is independent. To capture cross-frequency interactions efficiently without quadratic cost, we add a rank-R coupling term:
\begin{align} \label{rang_R}
Y^{(g)}(\omega) = D_g(\omega)X^{(g)}(\omega) + \lambda(t)\alpha_g\sum_{r=1}^RU_{g,r}(\omega)<V_{g,r}, X^{(g)}>,
\end{align}
where $g\in\{1,...,G\}$ indexes channel groups with $C_g=C/G$ channels per group, $R\in\mathbb{N}$ is the low-rank number, $\alpha_g\in\mathbb{R}$ is a learned coupling scale for group $g$, $\lambda(t)\in[0, 1]$ is a warmup factor at optimization step $t$, $U_{g,r}(\omega), V_{g,r}(\omega)\in\mathbb{C}$ are learned complex frequency basis functions, and the global spectral projection is:
\begin{align} \label{spectral_projection}
<V_{g,r}, X^{(g)}>=\sum_{\omega'\in\Omega^+}\overline{V_{g,r}(\omega')}X^{(g)}(\omega') \in \mathbb{R}^{B\times C_g},
\end{align}
where $\Omega^+$ denotes the half-spectrum frequency index set with $\rm L=|\Omega^+|=H\cdot W_f$.
This term implements projection and broadcast, which compresses the full spectrum into 
$R$ global coefficients and redistributes them across bins, yielding a non-diagonal operator of the form ${\rm Diag}(D)+\sum_{r=1}^Ru_rv_r^{H}$ with O(LR) computational complexity.

Since the full Fourier spectrum satisfies Hermitian symmetry, we apply projection $\Pi_{\mathcal{H}}$ onto the set of Hermitian-consistent half-spectra to ensure that the inverse transform produces a real-valued output and to suppress artifacts. Then, the final output of spectral branch can be expressed as:
\begin{align} \label{eq:Hermitian}
Y_{spe}={\rm IFFT}(\Pi_{\mathcal{H}}(Y(\omega))\odot {\rm SiLU}(F_{LN}).
\end{align}
The final output of DMB is the summation of outputs of dual branches with a residual connection after a feature fusion block (FFB):
\begin{align} \label{eq:DMBoutput}
& Y_{out} = w_1 \cdot Y_{spa}+w_2 \cdot Y_{spe}+X, \\
&Y_{\rm DMB}=
 Conv_{1\times 1}({\rm GELU}(DWConv_{3\times 3}(Conv_{1\times1}(Y_{out})))),
\end{align}
where $w_1$ and $w_2$ are learnable weights.

\begin{figure}[tb]
  \centering
  \includegraphics[width=.75\linewidth]{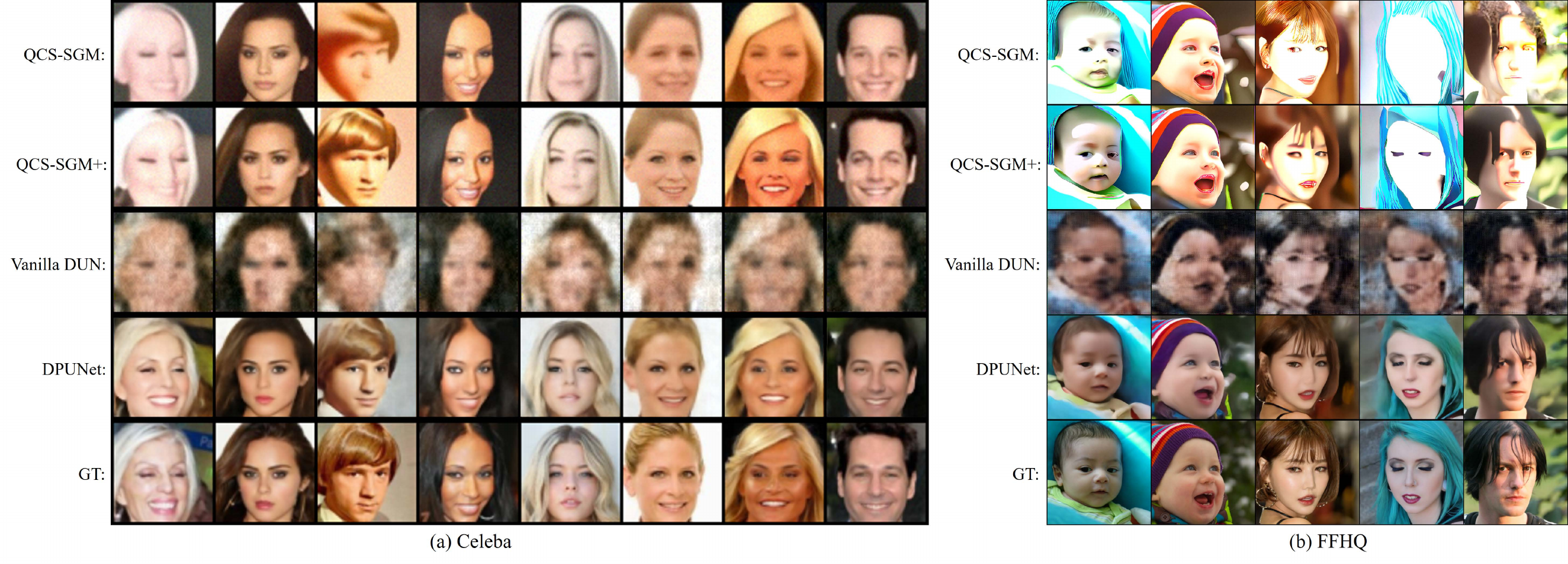}
  \caption{The reconstruction results from 1-bit measurements of CelebA dataset ($64\times 64$, the sampling number is 4000).
  }
  \label{fig:1bitceleba}
\end{figure}

\subsection{Loss Function}
We optimize the final output $\Xmat_K$ by minimizing the negative log-likelihood (NLL) of the quantized observations under the noise-perturbed model, which can be expressed as:
\begin{align} \label{eq:NLL}
\mathcal{L}_{\rm NLL}=\frac{1}{|\Omega_m|}\sum_{i\in\Omega_m}(-{\rm log}p(y|z_i)),
\end{align}
where $p(y|z_i)$ follow Eq.~\eqref{eq:cfd} and Eq.~\eqref{eq:1bitcasep} for multi-bit and 1-bit cases, respectively.
Thus, the final objective of the proposed DPUNet is:
\begin{align} \label{eq:}
\mathcal{L}=||X_K-X_{gt}||_2 + \alpha\mathcal{L}_{\rm NLL},
\end{align}
where $\alpha$ is empirically set to 0.05 in experiment.

\section{Simulation Results}
\begin{figure}[tb]
  \centering
  \includegraphics[width=.6\linewidth]{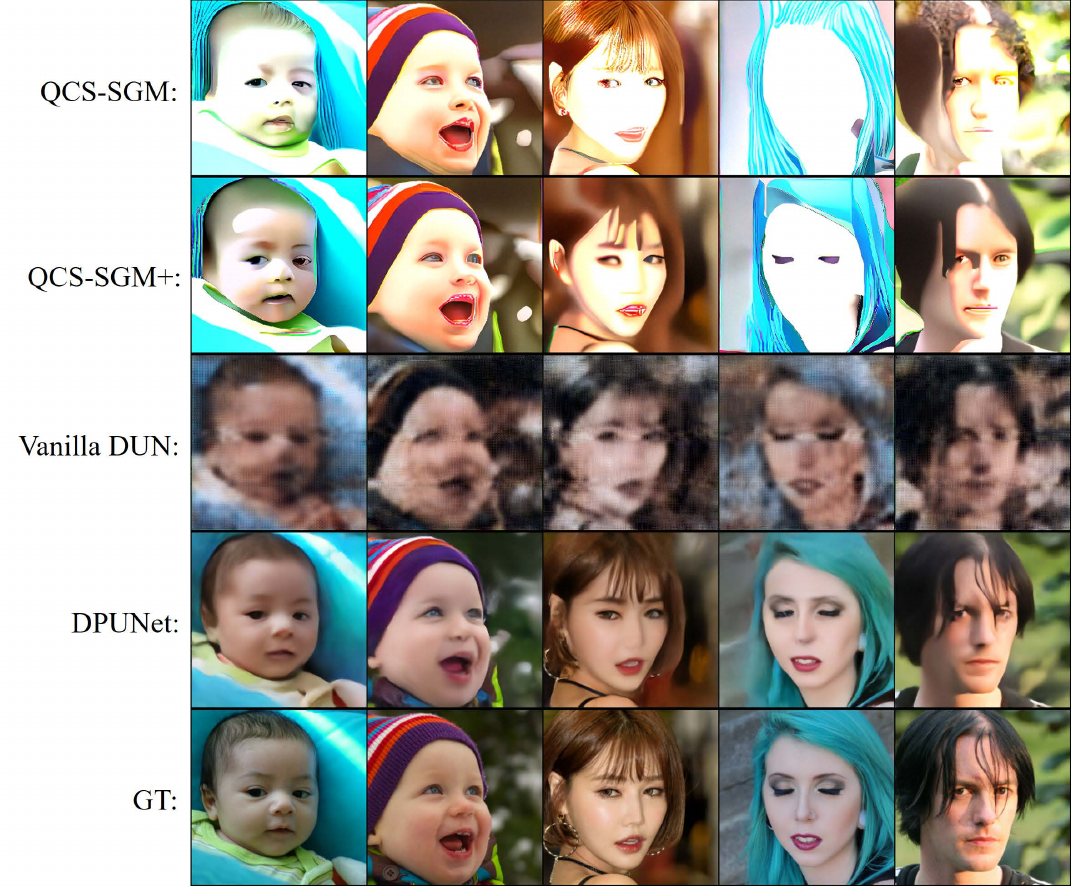}
  \caption{The reconstruction results from 1-bit measurements of FFHQ dataset ($256\times 256 \times 3$, the sampling number is 24576).
  }
  \label{fig:1bitffhq}
\end{figure}

\begin{figure}[tb]
  \centering
  \includegraphics[width=1\linewidth]{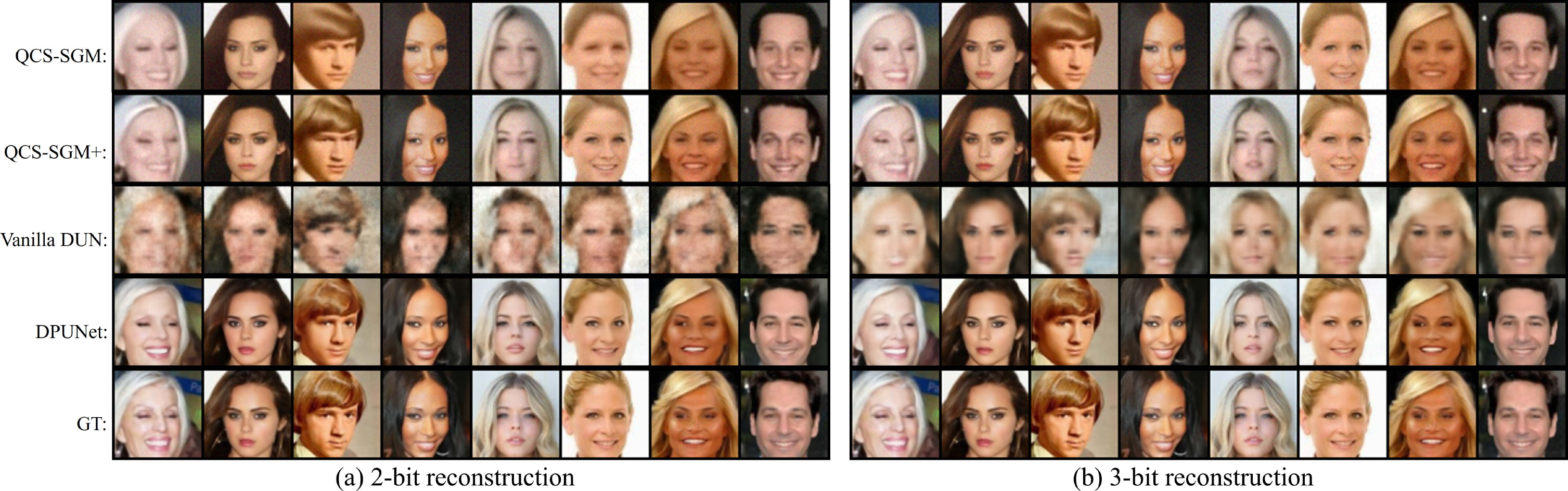}
  \caption{The reconstruction results of CelebA dataset from multi-bit measurements ($64\times 64 \times 3$, the sampling number is 4000).
  }
  \label{fig:C23}
\end{figure}

\begin{figure}[tb]
  \centering
  \includegraphics[width=.85\linewidth]{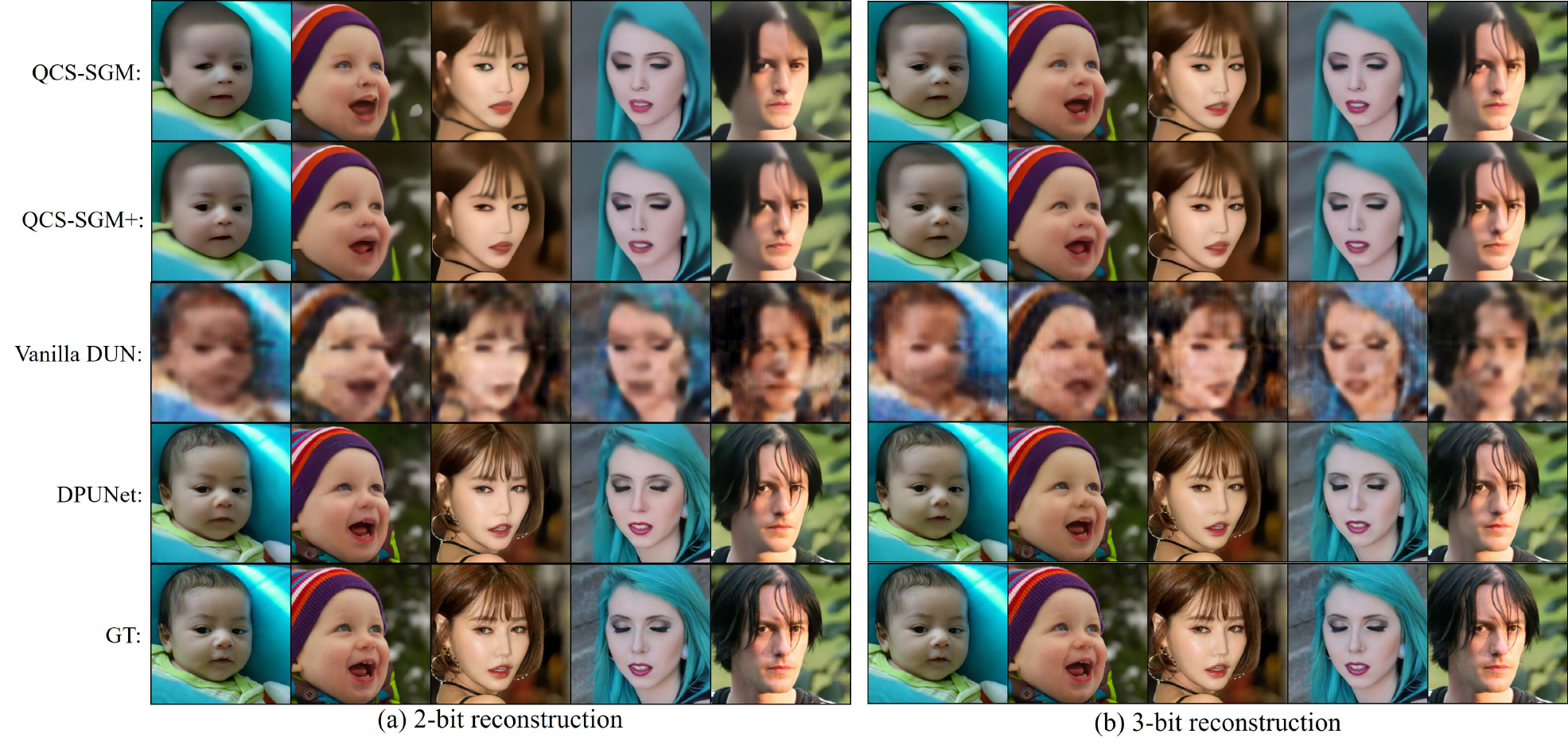}
  \caption{The reconstruction results of FFHQ dataset from multi-bit measurements  ($256\times 256\times 3$, the sampling number is 24576).
  }
  \label{fig:F23}
\end{figure}

We make a comprehensive comparison with previous SOTA methods to evaluate the performance of proposed DPUNet, including QCS-SGM~\cite{meng2023deep} and QCS-SGM+~\cite{meng2024qcs}. We further train a vanilla unfolding model, which simply replaces the likelihood data-consistency projection in DPUNet with the common proximal gradient descent projection in linear case~\cite{liao2025using, Wang_2025_CVPR, qu2024dual} to make a further comparison.
During training, we use 10,000 images from CelebA~\cite{liu2015deep} ($64\times64\times3$) and FFHQ~\cite{karras2017progressive} ($256\times256\times3$) as the training datasets, respectively. 50 images are randomly selected from the each dataset before training serve as the test datasets. 
Tab.~\ref{tab: simu} presents the average PSNR/SSIM and inference time of different methods under different quantization bits. Clearly, the proposed DPUNet outperforms previous methods at all QBs. 
Fig.~(\ref{fig:1bitceleba}) and Fig.~(\ref{fig:1bitffhq}) present the visualization comparison on the reconstruction results from 1-bit measurement. Fig.~\eqref{fig:C23} and Fig.~\eqref{fig:F23} present the comparison of reconstruction from 2-bit and 3-bit measurement. The proposed method showcases the best results and also improvements in details and textures under all QBs.

\begin{table*}[!htbp]
\centering
\caption{Average PSNR/SSIM and inference time of different methods on FFHQ dataset ($256\times 256\times3$, sampling number is 24576) and CelebA dataset ($64 \times 64 \times 3$, sampling number is 4000) with different quantization bits. The best and second best results are highlighted in \textbf{bold} and \underline{underlined}, respectively.}
\label{tab: simu}
\renewcommand\tabcolsep{6pt}
\scalebox{0.7}{
\begin{tabular}{c|c|ccc|c} 
\toprule[1pt]
\multicolumn{1}{c|}{\multirow{2}{*}{Dataset}}  &
\multicolumn{1}{c|}{\multirow{2}{*}{Method}}   & 
\multicolumn{3}{c|}{Quantization Bit (QB)} &  
\multirow{2}{*}{Inference Time}  \\ 
\cline{3-5}  
\multicolumn{1}{c|}{} &  
\multicolumn{1}{c|}{} &
\multicolumn{1}{c}{1 bit}  & 
\multicolumn{1}{c}{2 bit}  & 
\multicolumn{1}{c|}{3 bit} & \\ 
\toprule[1pt]
\multicolumn{1}{c|}{}  &
QCS-SGM &
{12.97}/{0.5147} & 
{24.27}/{0.6731} & 
{26.46}/{0.7265} &
{558 s} \\

\multicolumn{1}{c|}{FFHQ (256$\times$ 256 $\times$ 3)}  &
{QCS-SGM+}      & 
{13.26}/{0.5182} & 
\underline{25.25}/\underline{0.7034} &
\underline{27.62}/\underline{0.7838} &
1072 s \\
\multicolumn{1}{c|}{}  &
Vanilla DUN &
\underline{20.58}/\underline{0.6023} & 
{21.23}/{0.6628} & 
{22.89}/{0.7051} &
\textbf{0.22 s} \\
\multicolumn{1}{c|}{}  &
{DPUNet}      & 
\textbf{27.18}/\textbf{0.8481} & 
\textbf{32.51}/\textbf{0.9082} &
\textbf{33.46}/\textbf{0.9136} &
\underline{0.25 s} \\
\toprule[1pt]

\multicolumn{1}{c|}{}  &
QCS-SGM &
{15.42}/{0.6196} & 
{22.32}/{0.6628} & 
{24.24}/\underline{0.7853} &
{224 s} \\

\multicolumn{1}{c|}{CelebA (64$\times$ 64$\times$ 3)}  &
{QCS-SGM+}      & 
{17.27}/\underline{0.6697} & 
\underline{24.30}/{0.7828} &
\underline{25.82}/\underline{0.8083} &
653 s \\
\multicolumn{1}{c|}{}  &
Vanilla DUN &
\underline{20.69}/{0.6520} & 
{21.15}/{0.6757} & 
{22.66}/{0.7115} &
\textbf{0.11 s} \\
\multicolumn{1}{c|}{}  &
{DPUNet}      & 
\textbf{27.08}/\textbf{0.8990} & 
\textbf{31.53}/\textbf{0.9357} &
\textbf{32.54}/\textbf{0.9422} &
\underline{0.13 s} \\
\toprule[1pt]
\end{tabular}
}
\vspace{-3mm}
\end{table*}

\section{Ablation Study and Analysis.} 
\label{tab: ablation}
\subsection{Ablation Study}
To quantitatively analyze the effect of different components in DPUNet, we perform ablation study on FFHQ dataset at sampling ratio of $12.5\%$ and 3-bit quantization. The proposed DPUNet is mainly powered by the following designs: likelihood gradient projection (LGP), spatial branch and spectral branch in DMB, rank-R coupling in spectral branch.
The average PSNR and SSIM are shown in Tab.~\eqref{tab: ablation}. 
Baseline model (a) contains all components and presents the best result of 33.46 dB/0.9136. 
Towards model (b) without LGP, there is an
average 10.57 dB/0.2085 degradation on PSNR and SSIM. The absence of LGP in DPUNet leads to the most severe degradation in reconstruction, demonstrating the effectiveness of the proposed likelihood projection.
Towards model (c) without spatial branch in DMB, there is an
average 1.04 dB/0.0216 degradation on PSNR and SSIM.
Towards model (d) without spectral branch in DMB, there is an
average 2.17 dB/0.0874 degradation on PSNR and SSIM, meaning that the proposed $S^3M$, as a global perception module, is effective for QCS reconstruction.
Model (e) without  rank-R coupling in spectral branch, there is an
average 0.21 dB /0.0012 degradation on PSNR and SSIM, which demonstrates the importance of cross-frequency interactions in spectral domain.
\begin{table}[th]
\centering
\caption{Ablation study of the proposed DPUNet.} 
\renewcommand\tabcolsep{6.5pt}
\scalebox{0.80}{
\begin{tabular}{|c|c|c|c|c|c|c|}
\hline
Model & LGP & Spatial Branch & spectral branch & Rank-R coupling  & PSNR (dB) & SSIM   \\ \hline
(a) & \checkmark  & \checkmark  & \checkmark & \checkmark  & {\bf 33.46} & {\bf 0.9136} \\  \hline 
(b) &  &  \checkmark     &    \checkmark    & \checkmark  & 22.89 & 0.7051 \\ \hline
(c) & \checkmark  &  & \checkmark & \checkmark & 32.42 & 0.8920 \\ \hline
(d) & \checkmark  & \checkmark  &  &  \checkmark & 31.29 & 0.8262 \\  \hline
(e) & \checkmark &  \checkmark   &  \checkmark  &   & 33.25 & 0.9124 \\ \hline
\end{tabular}
}

\label{tab: ablation}
\end{table} 

\begin{figure}[tb]
  \centering
  \includegraphics[width=1\linewidth]{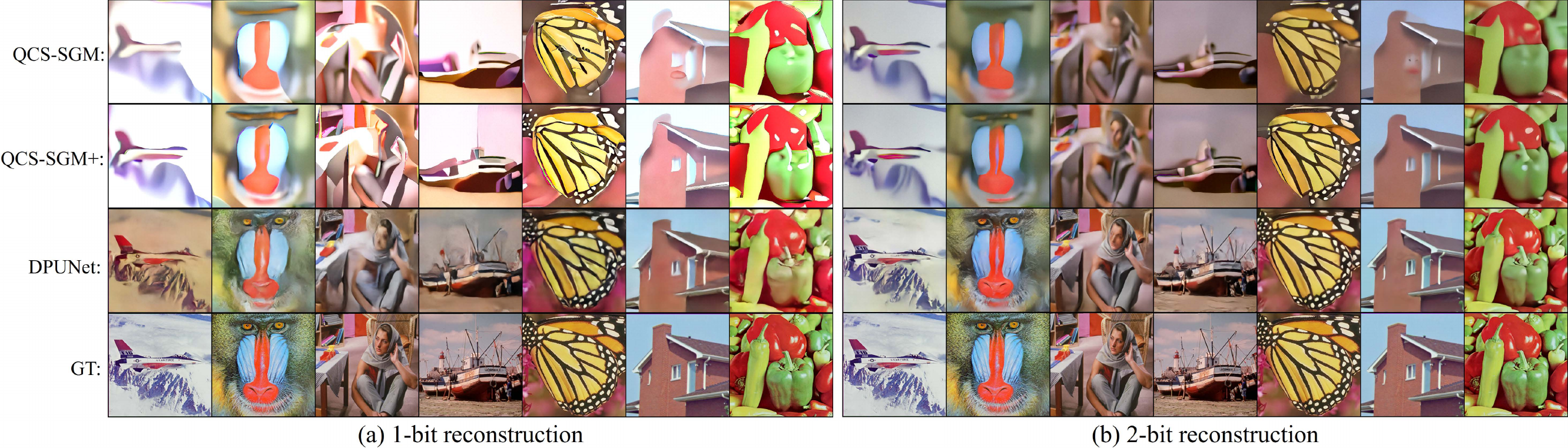}
  \caption{The OOD reconstruction results of CSet8 dataset from one- and two-bit measurements ($256\times 256\times 3$, the sampling number is 24576).
  }
  \label{fig:OOD}
\end{figure}

\subsection{Comparison of Computational Complexity.}
Then, we further make a comparison on the average PSNR, the number of parameters, FLOPs, and inference time with previous unfolding designs (image size is $256 \times\ 256 \times 3$) of different methods, as presented in Tab.~\ref{tab: inference}. Benefiting from the efficient design of $S^3M$ in dual-Mamba block, the proposed DPUNet demonstrates significant improvement over previous SOTA DUN models on efficiency.

\begin{table}[th]
\centering
\caption{Comparisons on parameters, FLOPs, and inference time for $256\times 256$ reconstruction of different unfolding methods at SR=$12.5\%$.} 
\renewcommand\tabcolsep{4.8pt}
\scalebox{0.85}{
\begin{tabular}{|c|c|c|c|c|c|}
\hline
Methods & CPPNet~\cite{guo2024cpp}     & HATNet~\cite{qu2024dual}    &  ProxUnroll~\cite{Wang_2025_CVPR} & DPUNet    \\ 
\hline
FLOPs (G)   & {153.47}  & 494.42   & \underline {107.73}  & {\bf 30.76} \\ 
\hline
Params (M)  & 12.31     & 31.28    & \underline{3.90}    & {\bf 2.91}  \\
\hline
InferenceTime (s) & 0.41  & 0.60 & \underline{0.27} & {\bf 0.25}  \\ 
\hline
\end{tabular}}
\label{tab: inference}
\end{table}
\vspace{-9mm}
\subsection{The Performance of Out-of-distribution.}
We further make a comparison on the different methods to reconstruct the out-of-distribution (OOD) dataset under different QBs, as shown in fig.~\ref{fig:OOD}. We select images from CSet8 dataset. The pretrained score network for QCS-SGM and QCS-SGM+, and the proposed DPUNet are all trained on the same FFHQ dataset. We can find that the proposed DPUNet preserves sufficient information even under 1-bit case and presents the best reconstruction results compared to the previous works.
\section{Conclusion}
In this paper, we propose DPUNet, a deterministic deep probabilistic unfolding framework for QCS reconstruction that compiles Bayesian data-consistency into an end-to-end trainable network, avoiding expensive posterior sampling while retaining principled likelihood guidance. To enhance representation power under severe information loss, we further introduce a dual-domain Mamba block that fuses spatial state-space modeling with an order-invariant, frequency-conditioned spectral operator augmented by efficient low-rank cross-frequency coupling, enabling multi-scale local–global feature interaction at low cost. Extensive experiments show that DPUNet achieves state-of-the-art performance with orders-of-magnitude faster inference than score-based sampling baselines across quantization bits, and it is also lightweight in parameters/FLOPs and runtime compared with prior unfolding designs, while exhibiting promising out-of-distribution reconstruction behavior. The proposed method could be a prototype for QCS reconstruction and has the potential to extend toward task-aware or hardware-in-the-loop optimization, where reconstruction is jointly tuned for downstream objectives, such as recognition or secure transmission.


%
%
\bibliographystyle{splncs04}
\bibliography{main}
\end{document}